%% file: 1st_draft.tex
\documentclass[graybox]{svmult}

\usepackage{type1cm}
\usepackage{makeidx}
\usepackage{graphicx}
\usepackage{multicol}
\usepackage[bottom]{footmisc}
\usepackage{hyperref}
\usepackage{newtxtext}
\usepackage{newtxmath}
\usepackage{inputenc}
\usepackage{graphicx}
\usepackage{filecontents}
\usepackage{multicol}
\usepackage{epstopdf}

\begin{document}

\title*{Optimal EEG Electrode Set for Emotion Recognition From Brain Signals: An Empirical Quest}
\titlerunning{Optimal EEG Electrode Set for Emotion Recognition From Brain Signals}

\author{Rumman Ahmed Prodhan, Sumya Akter, *Tanmoy Sarkar Pias, Md. Akhtaruzzaman Adnan}
\authorrunning{R. A. Prodhan et al.}

\institute{Rumman Ahmed Prodhan \at University of Asia Pacific, Dhaka, Bangladesh, \email{rumman153@gmail.com}
\and Sumya Akter \at University of Asia Pacific, Dhaka, Bangladesh, \email{sumyaakter601@gmail.com}
\and Tanmoy Sarkar Pias \at Virginia Tech, Blacksburg, VA 24061, United States, \email{tanmoysarkar@vt.edu} \\ {*Corresponding Author}
\and Md. Akhtaruzzaman Adnan \at University of Asia Pacific, Dhaka, Bangladesh, \email{adnan.cse@uap-bd.edu}}

\maketitle

\abstract{The human brain is a complex organ, still completely undiscovered, that controls almost all the parts of the body. Apart from survival, the human brain stimulates emotions. Recent research indicates that brain signals can be very effective for emotion recognition. However, which parts of the brain exhibit most of the emotions is still under-explored. In this study, we empirically analyze the contribution of each part of the brain in exhibiting emotions. We use the DEAP dataset to find the most optimal electrode set which eventually leads to the effective brain part associated with emotions. We use Fast Fourier Transformation for effective feature extraction and a 1D-CNN with residual connection for classification. Though 32 electrodes from the DEAP dataset got an accuracy of 97.34\%, only 12 electrodes (F7, P8, O1, F8, C4, T7, PO3, Fp1, Fp2, O2, P3, and Fz) achieve 95.81\% accuracy. This study also shows that adding more than 10 electrodes does not improve performance significantly. Moreover, the frontal lobe is the most important for recognizing emotion.}

\section{Introduction}
\label{sec:1}

On any given day, we experience a wide range of emotions. Some are happy, some are sad, and some are in between. But what are emotions, exactly? According to Psychology Today, "emotions are feelings caused by the body's automatic physical response to events." In other words, our emotions result from the biological stimuli triggered by certain events. For example, one might feel happy if they see someone they love. So, emotion is the mental state that arises spontaneously due to certain stimuli. It is a feeling accompanied by specific physiological changes in the body. Emotion can be positive or negative \cite{microstates} and can affect our thoughts, feelings, and behavior. Emotion recognition is the only interdisciplinary field that can combine psychology, computer science, neuroscience, and cognitive science. 
 
The ability to read and understand the emotions of others is crucial for effective communication. Interestingly, most emotions are expressed by facial expressions, which are interpreted subconsciously by the other person. According to Paul Ekman and Friesen's \cite{ekman1971} theory of emotion, there are six basic emotions: happiness, sadness, anger, fear, surprise, and disgust. While these expressions are universal, the intensity of each emotion can vary from culture to culture. The ability to identify and understand the emotions of others is a critical social skill. It enables us to navigate relationships, respond effectively to emotional cues, and empathize with others. Despite its importance, emotion recognition remains a challenging task. In order to better understand and recognize emotion, it is important to first have a general understanding of what emotion is. Emotion has been characterized as a complicated state of feeling that is related with thoughts, behaviors, and physiological changes. It is important to note that emotion is not just one feeling but a spectrum of feelings that can range from happiness and excitement to sadness and anger. There are various techniques to perceive emotions such as brain waves, facial expressions, body language, and tone of voice.
 
In this research, we have explored the effective ways to identify emotions from brain waves. EEG \cite{kumar2012} denotes electroencephalogram through which we can identify brain activity. We have used EEG to monitor the brain waves. It is a process by which we may examine a brain's electrical activity by putting tiny metal electrodes on the scalp. EEG has many uses \cite{mEEG,attention2020,Alzheimer2022}. Even some researchers are trying to decode imagined speech from EEG signals \cite{Imagined_speech}. This is why we need to know how electrodes monitor brain activity and provide us output as a person's mental state. So, it is necessary to know where to position the electrodes on the scalp that provide the correct emotion. Researchers have provided varied viewpoints on this and have designed several sets of electrodes to get optimal electrodes that can give a precise result. Decreasing the number of electrodes can correspondingly affect the accuracy of recognizing a pattern \cite{Tacke2022}. However, there is not any particular set of electrodes that delivers the best results. That is why it produces ambiguity among the researchers. Because we do not have any suitable answer concerning these electrode sets. 
 
In this study, we experimented with different types of electrode sets which are found in the literature and analyzed the results based on the performance. There are many EEG datasets are available like DEAP \cite{deap}, SEED \cite{seed}, AMIGOS \cite{AMIGOS}, MAHNOB-HCI \cite{MAHNOB}, and THINGS-EEG \cite{THINGS-EEG2022}. Among these, we have used the DEAP dataset for its popularity. DEAP dataset is constructed using a EEG cap containing 32 electrodes. We have conducted our experiment according to the sequence of DEAP dataset electrode rank. The cerebrum, cerebellum, and brainstem are the three primary areas of the brain. The cerebrum is the brain's biggest part. It is split into two half (left and right) and four lobes within each hemisphere (frontal, parietal, temporal, and occipital). The cerebrum regulates voluntary muscle movement, speech, thought, perception, reasoning, and emotion. We have also experimented with the most significant lobes to get accurate emotion.

Zhang et al. \cite{zhang2020} considered the brain regions for emotion creation and showed the use of just a modest number of electrodes positioned on the frontal area of the scalp. Goshvarpour et al. \cite{gosh2019} investigated the optimal electrode positions determined by the lagged Poincare's measurements of EEG recordings and a source localization approach.
 
Most of the previous research \cite{gosh2019,wang2019} utilized all 32 electrodes from the DEEP dataset. However, it is unlikely that all regions of the human brain would contribute to emotions. So, in this paper, we quest to find the brain regions associated with emotion by identifying an optimal set of electrodes. Some recent studies have proposed different strategies to acquire optimal electrodes for emotion recognition. They examined by applying their feature extraction approach together with chosen algorithms to acquire the best result. To best of our knowledge, there is no comparative analysis on effective feature extraction and machine learning algorithm for getting optimal electrode set for correct identification of emotion. So, we work to fill up this research gap by proposing a comparative analysis of different electrode sets and an optimal electrode set.  

The contribution in this work is to use our feature extraction approach, which is a fast Fourier transform \cite{FFT2011} to compare the electrode set we have found from literature to acquire the optimum electrode set along with the region of the electrode on the scalp.

The rest of paper is organized as follows: section 2 shows the previous related studies and their limitations; section 3 is dedicated for the DEAP dataset; section 4 demonstrates our methodology including preprocessing, feature extraction, and CNN modeling technique; section 5 demonstrates the experiment and result; and finally the study is concluded in section 6.

\begin{table}
\centering
\caption{Literature overview of optimal EEG channels used in DEAP dataset to recognize emotion}
\label{Literature overview of optimal EEG channels}       
%
%

\begin{tabular}{p{0.15\linewidth}p{0.09\linewidth}p{0.2\linewidth}p{0.13\linewidth}p{0.10\linewidth}p{0.15\linewidth}p{0.09\linewidth}p{0.11\linewidth}}
\hline\noalign{\smallskip}
Research &
Year &
Feature extraction &
Modeling technique & 
Number of 
electrodes &
Labels &
Accuracy\\
\noalign{\smallskip}\svhline\noalign{\smallskip}

Zhang et al. \cite{zhang2020}& 2020 & Wavelet entropy, Wavelet energy & KNN,NB,
SVM, RF & 12 & Valence, Arousal & 90\% \\
\\

Goshvarpour
et al. \cite{gosh2019}& 2019 & RSFS, SFFS, SFS & SVM & 05 & Valence, Arousal & 98.97\%, 98.94\% \\
\\

Joshi et al. \cite{joshi2020}& 2020 & Linear formulation of differential entropy, DE, Hjorth parmeter & biLSTM & 04 & 
Happy,
Angry,
Sad, Clam & 73.37\% \\
\\

Wang et al. \cite{wang2019}& 2019 & STFT  & SVM & 
08, 10 & Valence, Arousal & 74.41\%, 73.64\% \\
\\

Topic et al. \cite{topic2022}& 2022 & CGH & CNN, SVM & 10, 10 & Valence, Arousal, Dominance & 90.76\%, 92.92\%, 92.97\%  \\
\\

Msonda et al. \cite{ms}& 2021 & Wavelet Decomposition & AdaBoost, Second order polynomial, LR, SVC, RF & 08 & Valence & 90\% \\

\noalign{\smallskip}\hline\noalign{\smallskip}
\end{tabular}
\end{table}

\section{Literature Review}
\label{sec:2}

Numerous investigations for finding optimal electrode sets for detecting emotion have been conducted in recent years. Most researchers used the DEAP dataset, which is publicly available. In the DEAP dataset, emotions are classified into five labels which are valence, arousal, dominance, liking, and familiarity. In this study, we have used FFT for feature extraction. Many other feature extraction techniques are used in this literature. Choosing an optimal electrode set is essential for effectively classifying emotions \cite{2D_CNN}. Table \ref{Literature overview of optimal EEG channels} shows the optimal electrode set according to the previous publications.

Zhang et al. \cite{zhang2020} found 12 electrodes as an optimal electrode set. They used Wavelet entropy and Wavelet energy as a feature extrusion method. Moreover, KNN, NB, SVM, and RF are used as different modeling techniques. They ranked the optimal electrode set using mRMR and ReliefF techniques \cite{Mazumder2015}. Also, they worked on two labels of the DEAP dataset and got an average of 90\% accuracy.

Goshvarpour et al. \cite{gosh2019} used RSFS, SFFS, and SFS as feature extraction techniques. Also, they used SVM as a binary classifier. After that, they ranked the electrode set using the sLORETA method \cite{sLORETA2002} and got five electrodes as an optimal electrode set. Finally, They got an accuracy of 98.97\%for arousal and 98.94\% for valence.

Joshi et al. \cite{joshi2020} got 73.37\% of accuracy by using Linear formulation of differential entropy, DE, and Hjorth parameter as feature extraction. Also, they used biLSTM as modeling technique. Then They found four electrodes as an optimal electrode set. Moreover, they classified the emotion as Happy, Angry, Sad, and Clam.

Wang et al. \cite{wang2019} Found eight electrodes for valence and ten electrodes for arousal as an optimal electrode set by using short-time Fourier transform as a feature extrusion method. Moreover, they used SVM as a modeling technique. After that, they ranked the optimal electrode set using the Normalized mutual information (NMI) connection matrix technique. Finally, they got an accuracy of 74.41\% for valence and 73.64\% for arousal from this optimal electrode set.

Topic et al. \cite{topic2022} got the accuracy of 90.76\% for valence, 92.92\% for arousal, and 92.97\% for dominance by using Computer-generated holography (CGH) \cite{CGH2000} as a feature extraction method. Also, they used CNN and SVM as a classifier. After that, they ranked the electrode by using ReliefF and Neighborhood Component Analysis (NCA) method and found ten electrodes for ReliefF and ten electrodes for Neighborhood Component Analysis (NCA) as an optimal electrode set.

Lastly, using the Mean Squared Error(MSE) method, Msonda et al. \cite{ms} found eight electrodes as an optimal electrode set. They achieved 67\% of accuracy by using Wavelet Decomposition as a feature extraction technique. Also, they used different modeling techniques like AdaBoost, Logistic Regression, Linear Support Vector Classifier (SVC), second order polynomial, and Random Forest (RF).

\begin{figure}
\centering
\includegraphics[width=\textwidth]{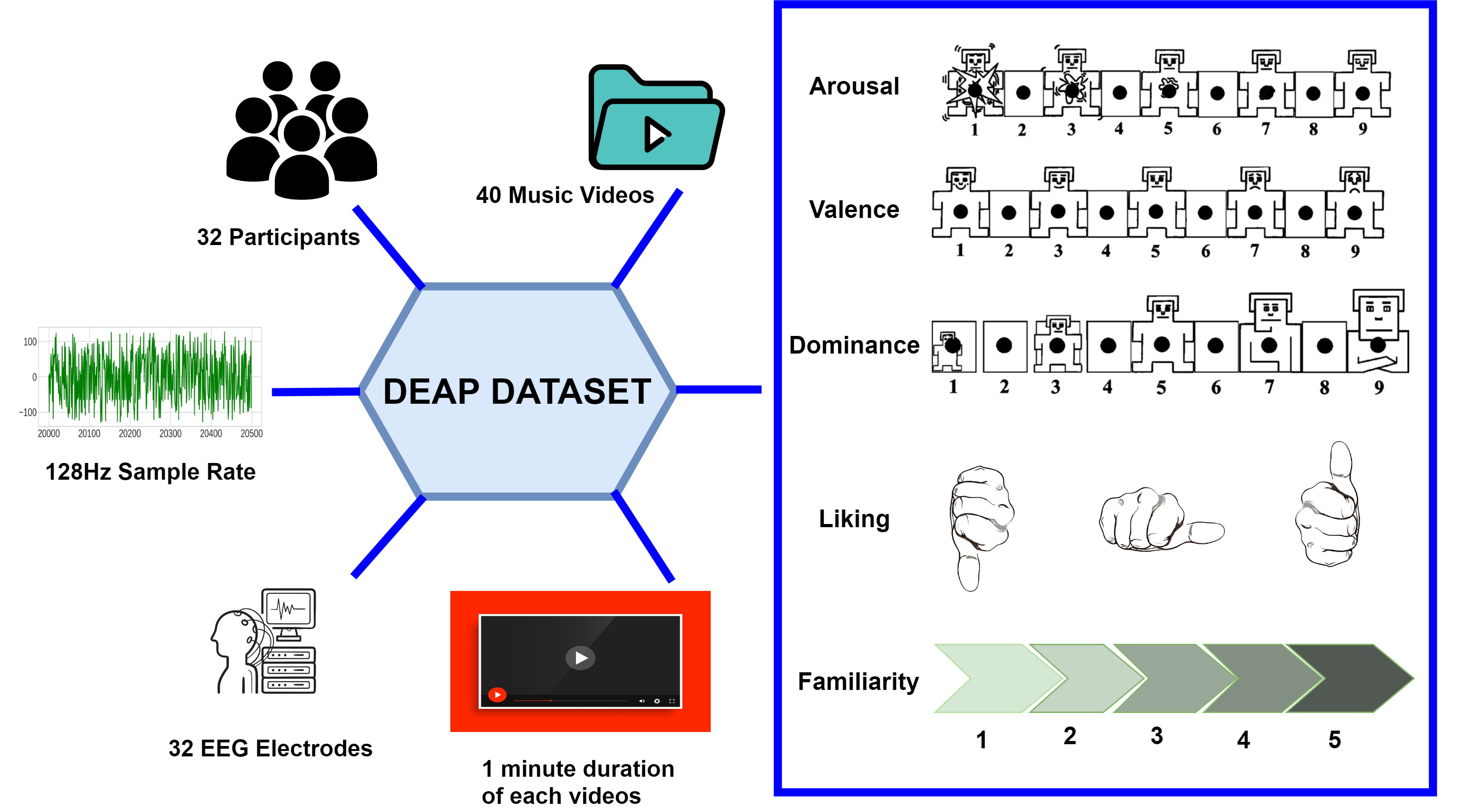}
\caption{Overview of the DEAP Dataset} 
\label{Overview of the DEAP Dataset}
\end{figure}

\section{Dataset}
\label{sec:3}

The DEAP dataset is multi-modal in nature, used to assess humans' emotional states. A group consists of 32 participants were shown 120 music videos with a one-minute length. Each participant has a .bdf file in the DEAP dataset, and they are all recorded with 48 channels at 512Hz. Two distinct locations have been used to record this dataset. Participants 1-22's data are logged in Twente, whereas participants 23-32's are logged in Geneva. The summary of the DEAP dataset is provided in figure \ref{Overview of the DEAP Dataset}.

In the DEAP dataset, additional preprocessed files have been provided, which are down-scaled to 128Hz. The DEAP dataset offers 2 down-sampled zip files. The preprocessed Python zip file, which contains 32 files in .dat format, is used for this study. A single .dat file represents a single participant. There are 2 arrays in each file. The data array contains 8064 data over 40 channels across 40 trials. Every trial lasts for 63 seconds. As a result, the data is 128x63=8064.

\begin{figure}
\includegraphics[width=\textwidth]{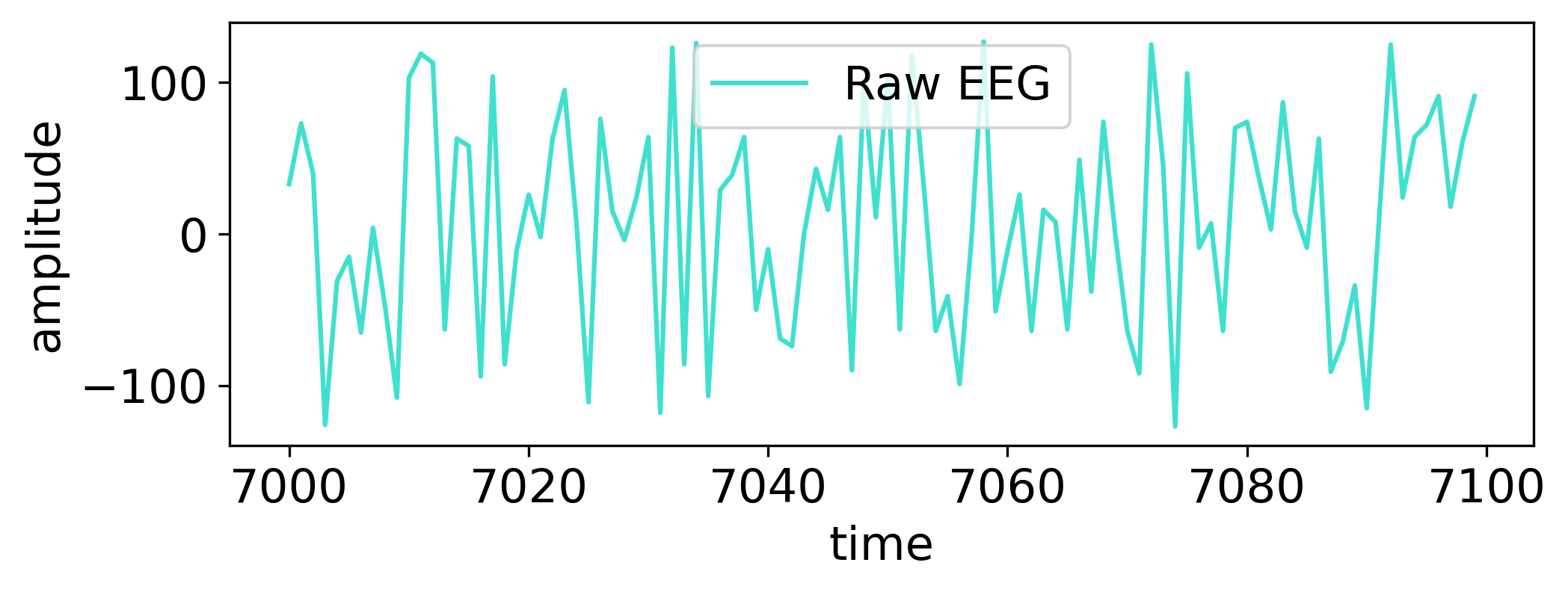}
\caption{A subject's raw EEG data-segment from the DEAP dataset} \label{Individual subject's raw EEG from the DEAP dataset}
\end{figure}

A participant's .dat file from the preprocessed Python folder is plotted in figure \ref{Individual subject's raw EEG from the DEAP dataset}. Each participant's EEG signals look like this. The label array consists of four labels—valence, arousal, dominance, and liking—and 40 trials. Among the 40 channels there are 32 EEG channels. Also, there are 8 other channels.

\section{Methodology}

\begin{figure}
\centering
\includegraphics[width=\textwidth]{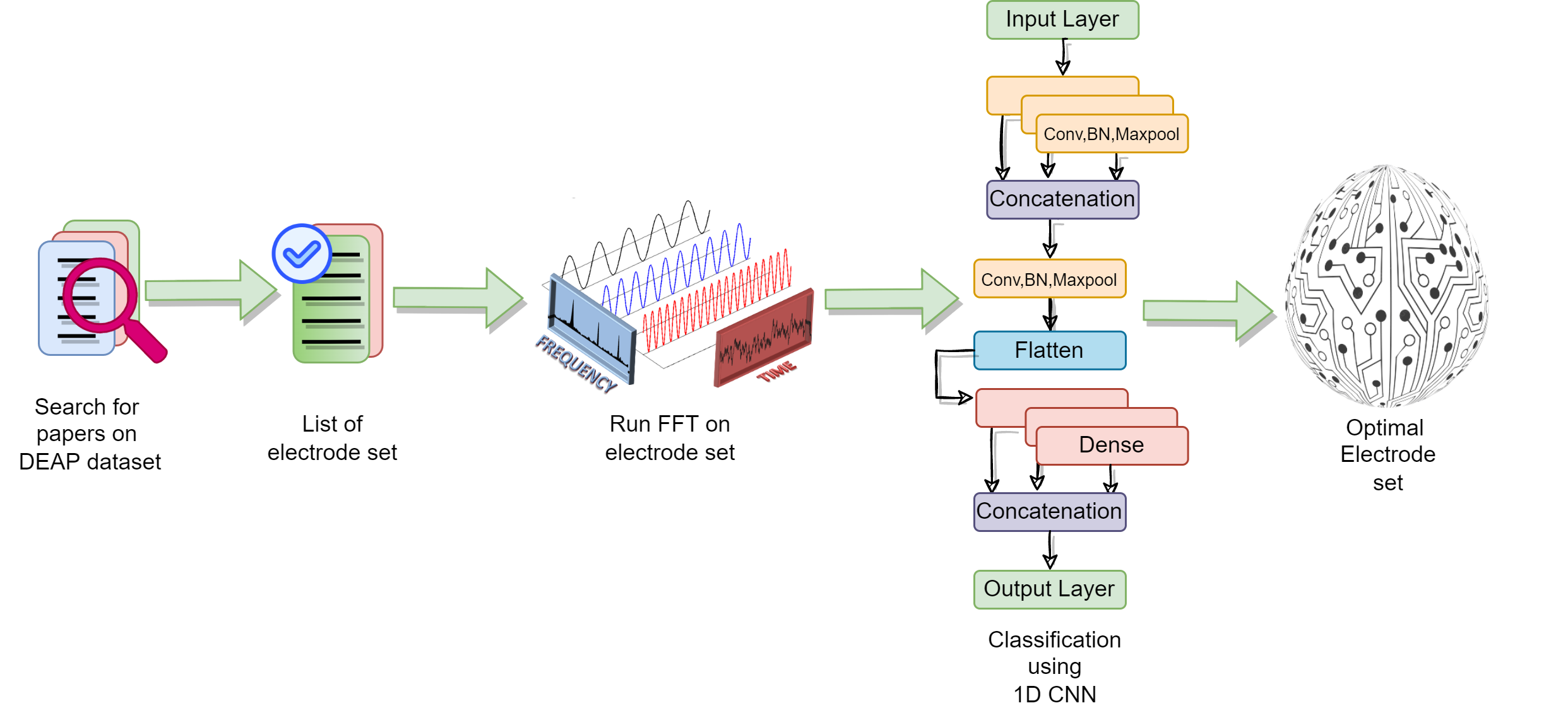}
\caption{Workflow diagram for searching for optimal electrode set to recognize emotions using 1D-CNN} \label{Complete workflow}
\end{figure}

Figure \ref{Complete workflow} illustrates the workflow of finding the optimal EEG electrodes for emotion recognition. At first, the DEAP dataset is selected as the source of raw EEG data. Then different types of preprocessing were applied to remove noise and artifacts. After that using FFT, the feature extraction is performed on the specific electrode set mentioned in the literature. All of the test was performed on a 1D CNN to maintain the consistency of the evaluation.

\subsection{Preprocessing}

The 512Hz EEG data has been downsized into 128Hz. Eye artifacts are eliminated using a blind source separation technique \cite{blind1997}. A bandpass frequency filter with 4.0-45.0Hz is implemented. The data is averaged following the commonly used reference. The EEG channels are rearranged following the Geneva order because the EEG data was recorded in two distinct places. Each trial's data is divided into 60 seconds and a baseline of 3 seconds. The pre-trial phase is then trimmed out. Additionally, the trials are rearranged to experiment video order instead of the presentation order.

\subsection{Feature Extraction}

\begin{figure}
\includegraphics[width=\textwidth]{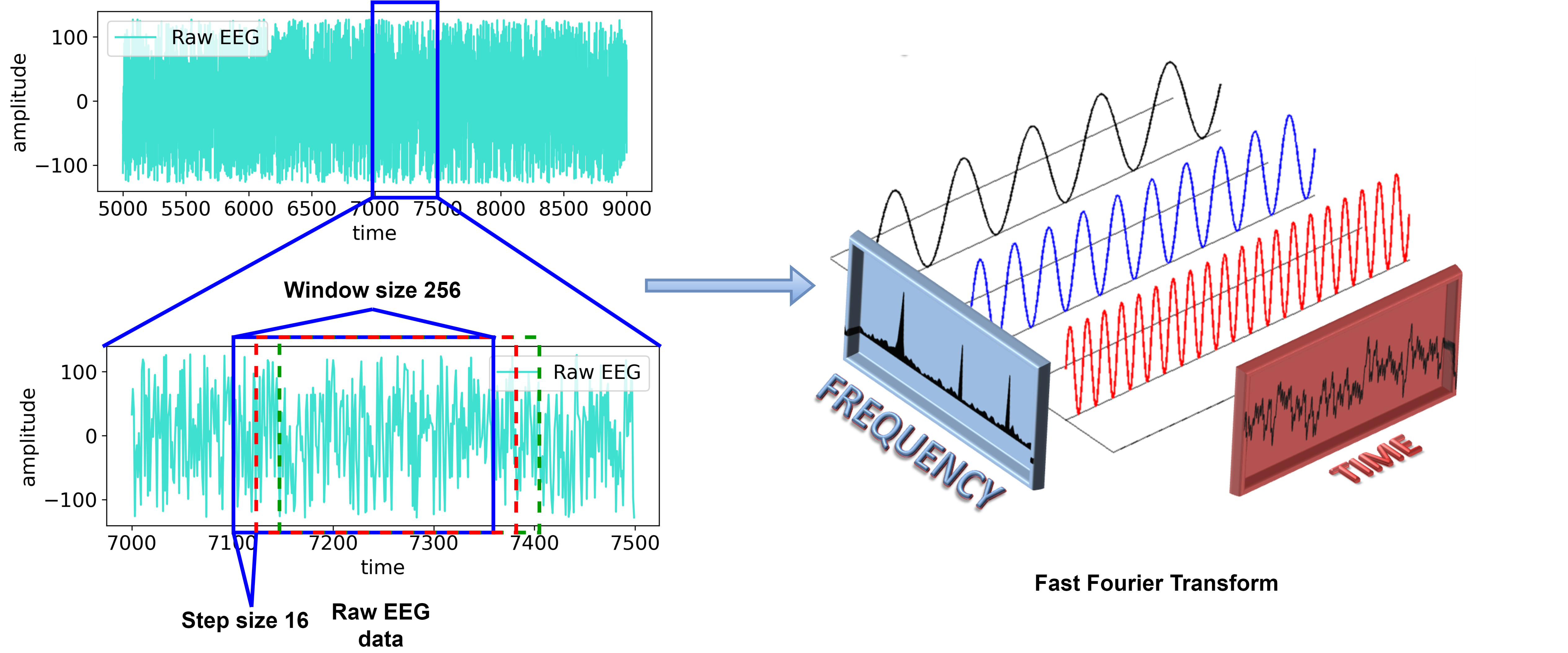}
\caption{Feature Extraction: FFT using sliding window} \label{Feature extraction using FFT}
\end{figure}

\begin{figure}
\centering
\includegraphics[width=\textwidth]{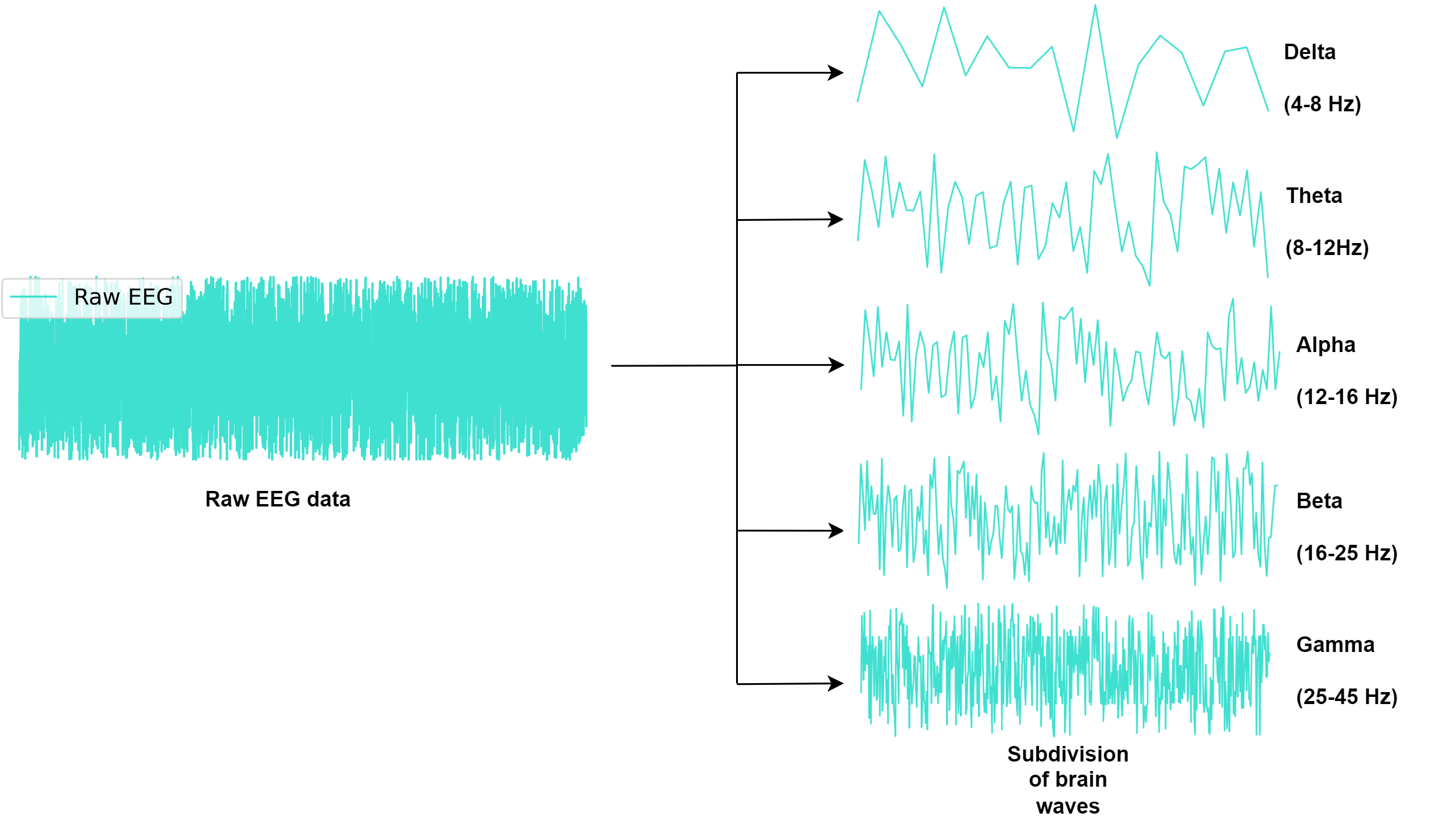}
\caption{Decomposing raw EEG data into five sub-bands} \label{Sub-bands}
\end{figure}

Feature extraction is critical for effective learning, minimizing signal loss, overfitting, and computational overhead. In general, designing an effective feature extraction method can produce better classification performance than raw data. Frequency-domain features are used to break down signal data into the subbands represented in figure \ref{Feature extraction using FFT}. we have used 256 as a window size and 16 as a step size in this experiment. Wavelet transform (WT), fast Fourier transform (FFT), equivocator methods (EM), etc., are frequently used for EEG feature extraction. Among these techniques, fast Fourier transform is proved to be the most effective according to recent publications \cite{hasan2021}; thus, this technique is used in this study. The discrete Fourier transform or inverse discrete Fourier transform (DFT) of a sequence can be determined using the FFT technique. Most of the actual signal is composed of several frequencies. The Fourier transformation is an effective method to extract those fundamental frequencies. After, decomposing the raw EEG data with a fast Fourier transform, we have found 5 sub bands of brain waves. These are Delta ranges from 4-8Hz, Theta ranges from 8-12Hz, Alpha ranges from 12-16Hz, Beta ranges from 16-25Hz, and Gamma ranges from 25-45Hz shows in figure  \ref{Sub-bands}. An actual signal can be long, so to make it faster and more accurate, FFT is used where the whole signal is divided into multiple segments on which FT is applied to extract frequencies.
The DTF can be expressed as follows:
\begin{equation}
x[k]=\sum_{n=0}^{N-1} x[n]e^\frac{-j2\pi kn}{N}
\end{equation}
In this case, the domain size is n. Each value of a discrete signal x[n] should be multiplied by an e-power to some function of n to determine the DFT for that signal. The results obtained for a given n should then be added together. Calculating a signal's DFT is O(N\textsuperscript{2}) in complexity. As its name suggests, fast Fourier Transform (FFT) is much quicker than discrete Fourier Transform (DFT). The complexity is reduced using FFT from O(N\textsuperscript{2}) to O (NlogN).

\subsection{EEG Electrodes Set}

\begin{figure}
\centering
\includegraphics[width=\textwidth]{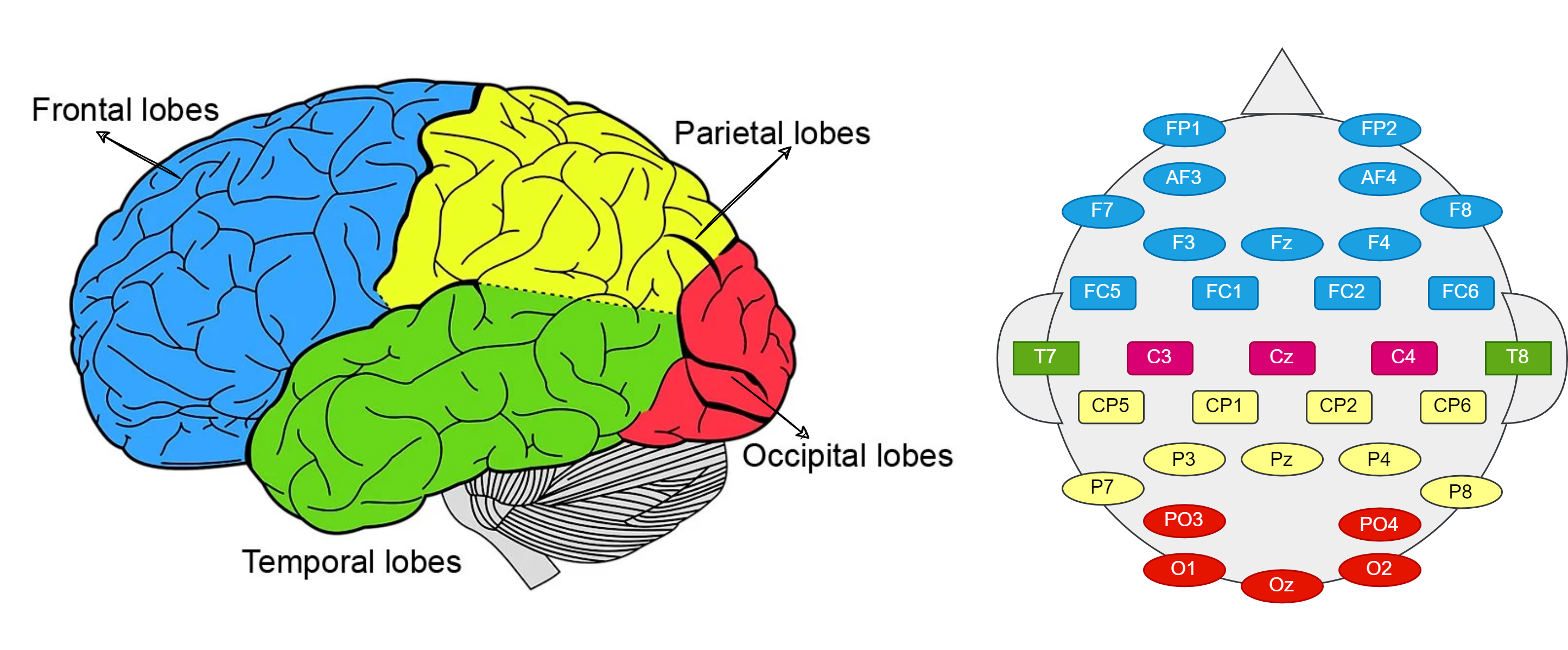}
\caption{Illustration of the four main lobes of cerebral hemisphere \cite{lobe2022} (left) and position of the 32 electrodes used in DEAP dataset on the scalp \cite{imagecite2022} (right)} \label{Position of the 32 electrodes used in DEAP dataset} \label{Position of the 32 electrodes used in DEAP dataset}
\end{figure}

The DEAP dataset team used an international 10-20 electrode placement system \cite{electrode1020} to collect the EEG signals. The majority of the EEG electrodes associated with emotions are the frontal lobe denoted by the color blue, the parietal lobe is yellow, the occipital lobe is red, the temporal lobe is green, and the center areas in squares shown in the left of figure \ref{Position of the 32 electrodes used in DEAP dataset}. The letters FP, AF, F, FC, T, P, and O stand for the front polar, anterior frontal, frontal, front central, temporal, parietal, and occipital regions of the brain. An odd number suffix represents the left hemisphere, whereas an even number suffix represents the right hemisphere. These regions perfectly mirror how emotions are created physiologically. By changing the electrode distribution, it is possible to decrease the extracted feature dimension. The experiment can be made simpler and carried out more efficiently by reducing the complexity of the calculations. The position of the 32 EEG electrodes on the scalp is shown in the right of figure \ref{Position of the 32 electrodes used in DEAP dataset}.

\begin{table}[!t]
\caption{Electrode mapping according to the lobes of cerebral hemisphere}
\label{Electrodes according to the brain areas}       
%
%
\centering
\begin{tabular}{p{2cm}p{3cm}p{6.3cm}}
\hline\noalign{\smallskip}
Brain Area & Number of Electrodes & Electrodes \\
\noalign{\smallskip}\svhline\noalign{\smallskip}
Frontal & 13 & Fp1, Fp2, AF3, AF4, F7, F8, F3, Fz, F4, FC5, FC1, FC2, FC6 \\ 
Parietal & 9 & CP5, CP1, CP2, CP6, P7, P3, Pz, P4, P8 \\
Occipital & 5 & PO3, PO4, O1, Oz, O2 \\
Temporal & 2 & T7, T8 \\
Central & 11 &FC5, FC1, FC2, FC6, C3, Cz, C4, CP5, CP1, CP2, CP6 \\
\noalign{\smallskip}\hline\noalign{\smallskip}
\end{tabular}
\end{table}

Firstly, identifying the part of the brain which are responsible for emotion recognition is essential. Table \ref{Electrodes according to the brain areas} shows the group of electrodes according to the brain areas. Those are mainly frontal, parietal, occipital, temporal, and central.

\begin{table}
\caption{Electrode sets proposed in different publications}
\label{Electrode sets by different authors}       
%
%
\begin{tabular}{p{1cm}p{2.5cm}p{2.5cm}p{5.3cm}}
\hline\noalign{\smallskip}
Set No. & Research & Ranking Method & Electrodes\\
\noalign{\smallskip}\svhline\noalign{\smallskip}
01 & Zhang et al. \cite{zhang2020} & mRMR & F7, P8, O1, F8, C4, T7, PO3, Fp1, Fp2, O2, P3, Fz \\
02 & Zhang et al. \cite{zhang2020} & ReliefF & PO3, F8, Fp1, P3, Fp2, F3, O2, P8, Oz, F7, T8, Cz \\
03 & Goshvarpour et al. \cite{gosh2019} & sLORETA & FP1, C3, Cp1, P3, Pz \\
04 & Joshi et al. \cite{joshi2020}& Prefrontal & FP1, AF3, FP2, AF4 \\
05 & Wang et al. \cite{wang2019} & NMI & FC1, P3, Pz, Oz, CP2, C4, F4, Fz \\
06 & Topic et al. \cite{topic2022}& ReliefF & FP1, AF3, F3, F7, T7, O1, OZ, FP2, F8, P8 \\
07 & Topic et al. \cite{topic2022} & NCA & FP1, AF3, F7, T7, CP5, P7, FP2, AF4, FC6, T8 \\
08 & Msonda et al. \cite{ms}& Mean Squared Error & CP6, F3, F8, Fp1, O2, P7, T7, T8 \\
09 & - & - & Fp1, AF3, F3, F7, FC5, FC1, C3, T7, CP5, CP1, P3, P7, PO3, O1, Oz, Pz, Fp2, AF4, Fz, F4, F8, FC6, FC2, Cz, C4, T8, CP6, CP2, P4, P8, PO4, O2 \\
\noalign{\smallskip}\hline\noalign{\smallskip}
\end{tabular}
\end{table}

After identifying the part of the brain which are responsible for emotion, it is important to find out the specific set of EEG electrodes. From the literature, the best work on emotion recognition along with their optimal EEG electrode set are extracted. All the electrodes are selected according to the valence label on the DEAP dataset to maintain consistency. The selected EEG electrode sets are shown in table \ref{Electrode sets by different authors}.

\subsection{CNN Model Structure}

\begin{figure}
\includegraphics[width=\textwidth]{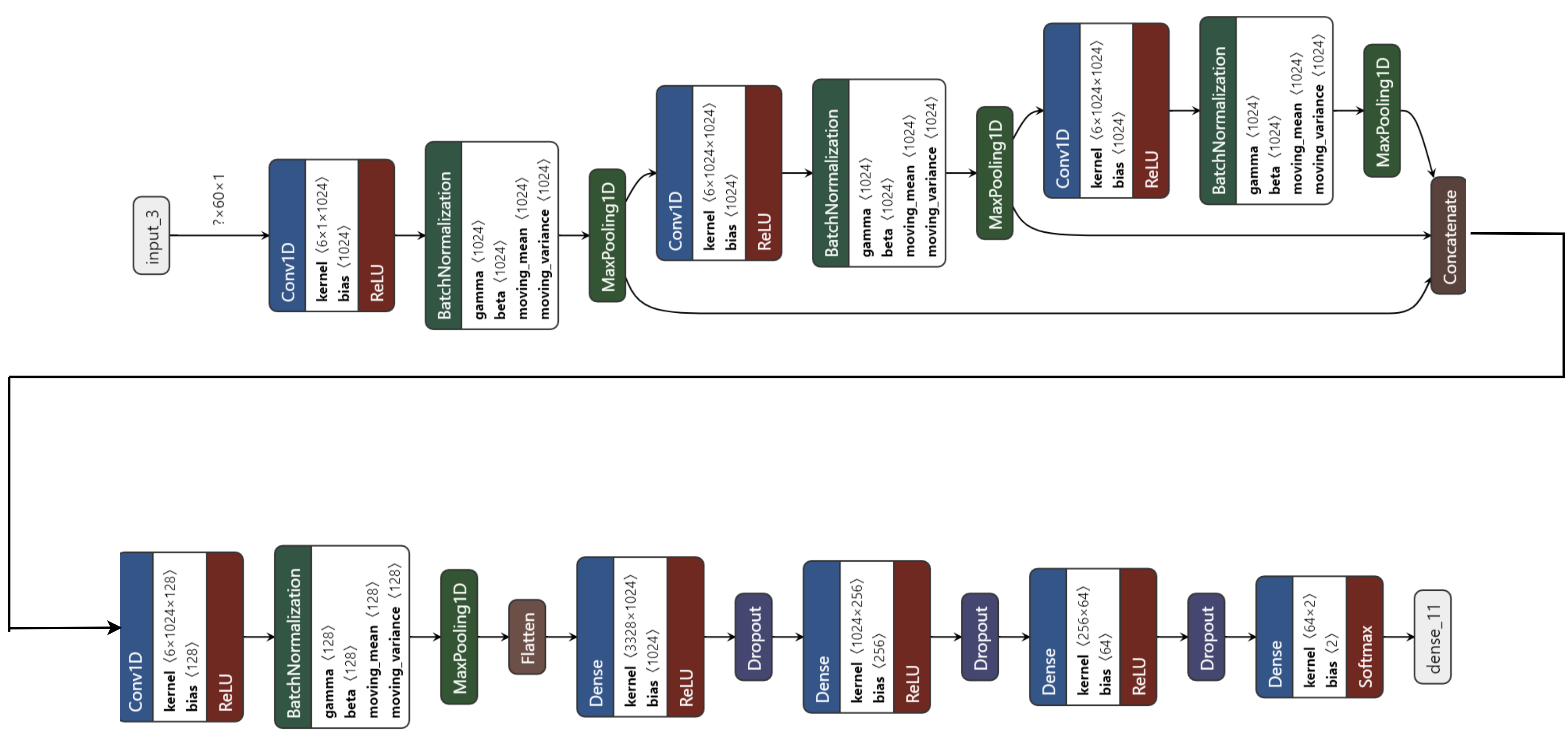}
\caption{Proposed 1D-CNN model architecture with residual connection} \label{Proposed CNN model architecture}
\end{figure}

Numerous signal processing tasks, such as early arrhythmia identification in electrocardiogram (ECG) beats \cite{ECG}, emotion recognition from EEG, activity recognition job from accelerometer data, etc., have made 1D CNN increasingly popular. Thus in this study, a 1D CNN with hidden layers shown in figure \ref{Proposed CNN model architecture} is used to accurately recognize emotions from brain signals. The proposed model uses the residual connection, a kind of skip connection \cite{skip2020}. Initially, the model is changed from sequential to functional to construct the residual connection. The input size of the model depends on the number of electrodes multiplied by the number of bands. Thus, the input size must be changed for each EEG electrode set.

\section{Experiment and Result Analysis}

Binary class classification on valence label is used for all the EEG sets to maintain the consistency of experiments. The same 1D CNN is used for all the sets. Residual connections are used to improve performance.

\begin{figure}
\centering
\includegraphics[width=\textwidth]{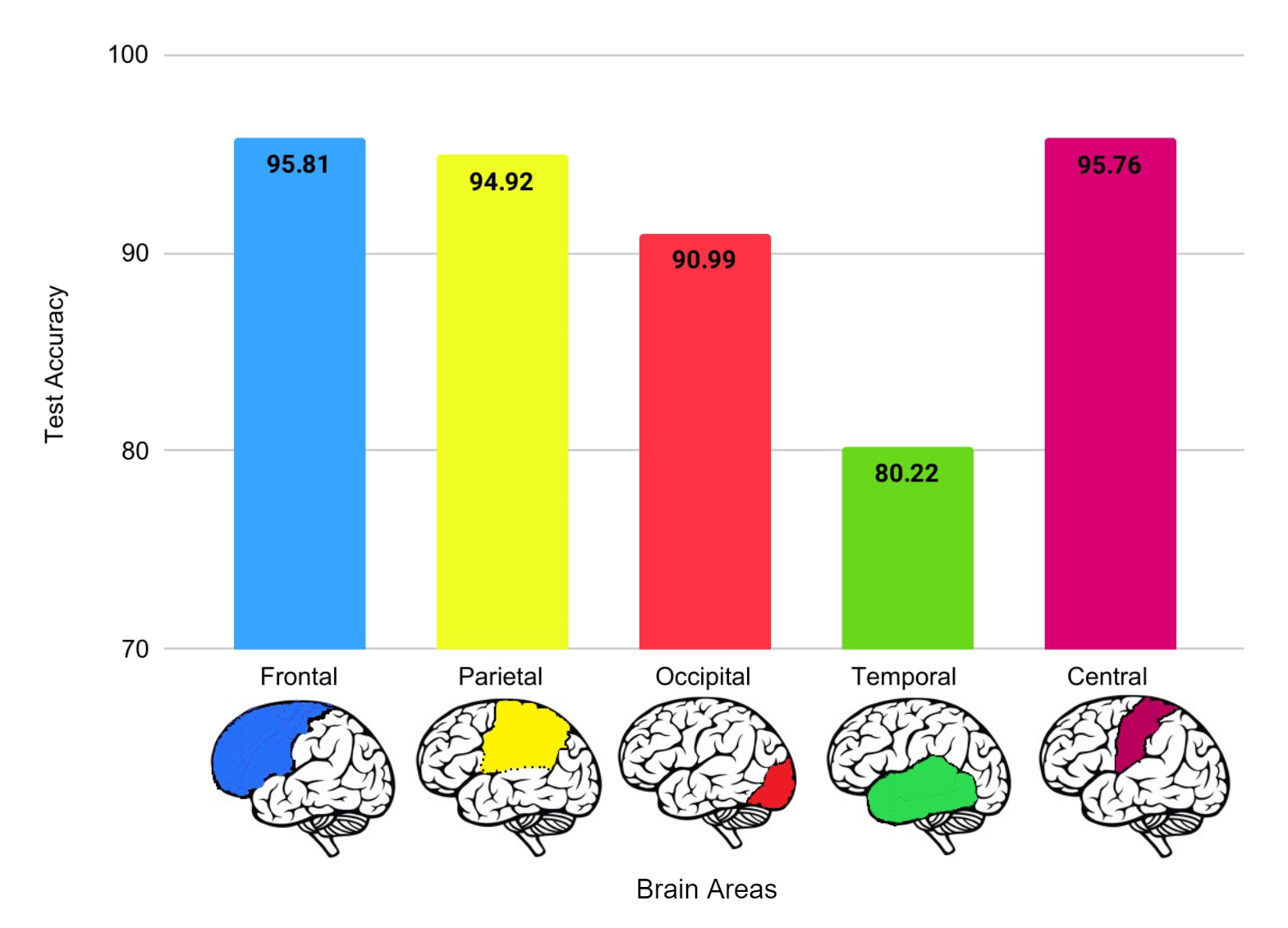}
\caption{Test accuracy of electrode sets according to the lobes of cerebral hemisphere} \label{accuracy of different parts of the brain}
\end{figure}

Figure \ref{accuracy of different parts of the brain}, illustrates the testing accuracy for the individual electrode sets according to the brain area. The frontal lobe's electrode sets perform best for emotion recognition. Following that, the central lobe is the second best. The frontal lobe is more critical for recognizing emotions than other brain areas. 

\begin{figure}
\centering
\includegraphics[width=\textwidth]{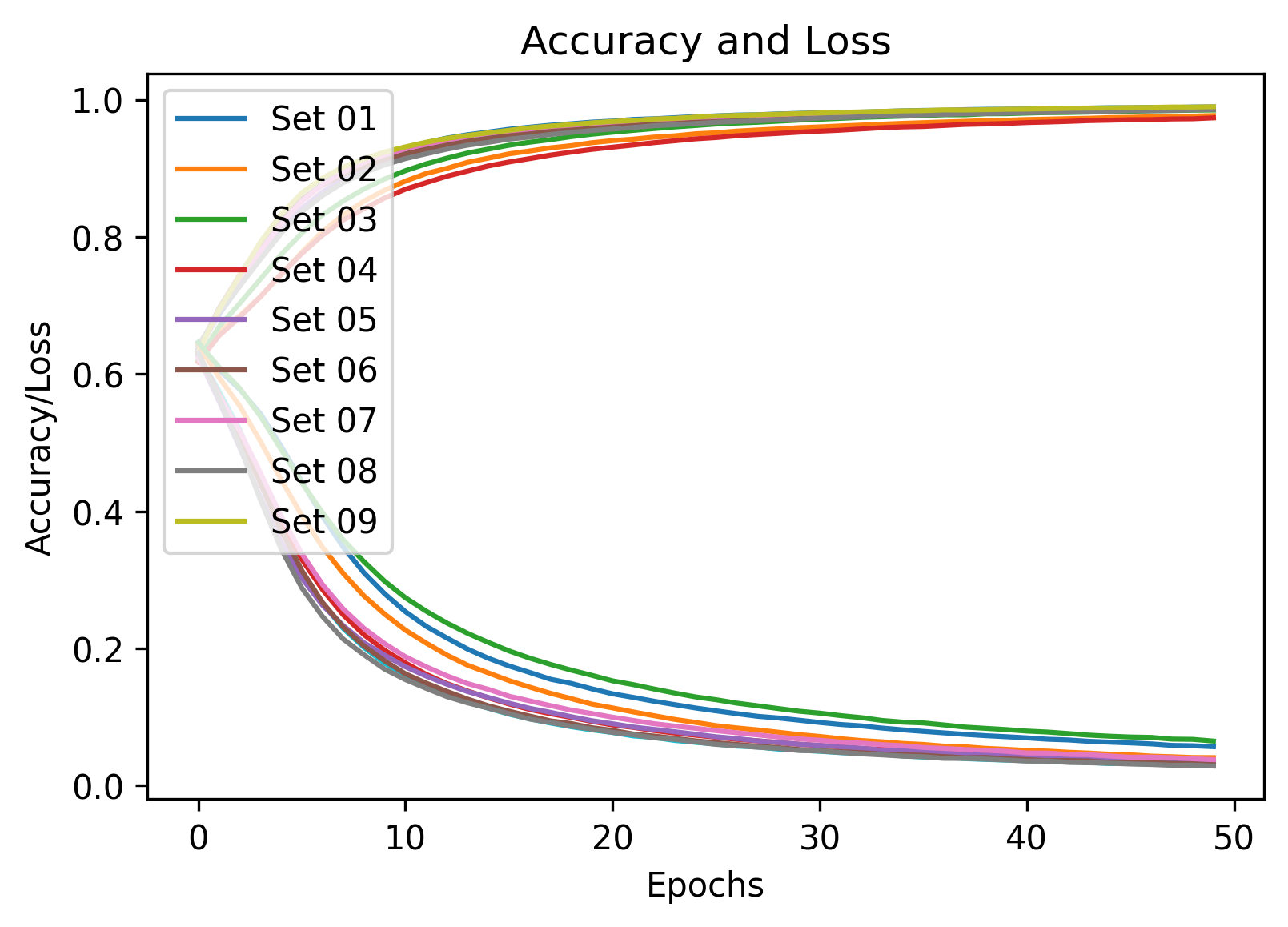}
\caption{Training accuracy and loss for two label classification on nine electrode sets} 
\label{Training accuracy and loss}
\end{figure}

Figure \ref{Training accuracy and loss} plots the training accuracy and loss of every epoch for all the nine electrode sets in table \ref{Electrode sets by different authors}. For the first 20 epochs, a massive increase in accuracy and decrease in loss is observed. The improvement is moderate from 20 to 30 epochs, and the progress is meagre from 30 to 50 epochs.

\begin{table}[!t]
\caption{Comparing results of electrode sets}
\label{Results of electrode sets by different authors}       
%
%
\begin{tabular}{p{6.3cm}p{2.5cm}p{2.5cm}}
\hline\noalign{\smallskip}
Electrodes & 
Previous accuracy & 
Our accuracy \\
\noalign{\smallskip}\svhline\noalign{\smallskip}

F7, P8, O1, F8, C4, T7, PO3, Fp1, Fp2, O2, P3, Fz & 90\% \cite{zhang2020}& \textbf{95.81\%} \\ 
\\
PO3, F8, Fp1, P3, Fp2, F3, O2, P8, Oz, F7, T8, Cz & 90\% \cite{zhang2020}& \textbf{95.60\%} \\ 
\\
FP1, C3, Cp1, P3, Pz & \textbf{98.97\%} \cite{gosh2019}& 93.63\%  \\ 
\\
FP1, AF3, FP2, AF4 & 73.37\% \cite{joshi2020}& \textbf{91.08\%} \\ 
\\
FC1, P3, Pz, Oz, CP2, C4, F4, Fz & 74.41\% \cite{wang2019}& \textbf{95.01\%} \\ 
\\
FP1, AF3, F3, F7, T7, O1, OZ, FP2, F8, P8 & 90.76\% \cite{topic2022}& \textbf{95.67\%} \\ 
\\
FP1, AF3, F7, T7, CP5, P7, FP2, AF4, FC6, T8 & 90.76\% \cite{topic2022}& \textbf{95.51\%} \\ 
\\
CP6, F3, F8, Fp1, O2, P7, T7, T8 & 90\% \cite{ms}& \textbf{94.11\%}  \\ 
\\
Fp1, AF3, F3, F7, FC5, FC1, C3, T7, CP5, CP1, P3, P7, PO3, O1, Oz, Pz, Fp2, AF4, Fz, F4, F8, FC6, FC2, Cz, C4, T8, CP6, CP2, P4, P8, PO4, O2 & 75.16\% \cite{wang2019}& \textbf{97.34\%}  \\ 

\noalign{\smallskip}\hline\noalign{\smallskip}
\end{tabular}
\end{table}

Table \ref{Results of electrode sets by different authors} shows the previous accuracy of each electrode set proposed by different publications and the accuracy achieved by the proposed model. All the electrode set is run with the CNN model shown in figure \ref{Proposed CNN model architecture} for 50 epochs.

\begin{figure}
\includegraphics[width=\textwidth]{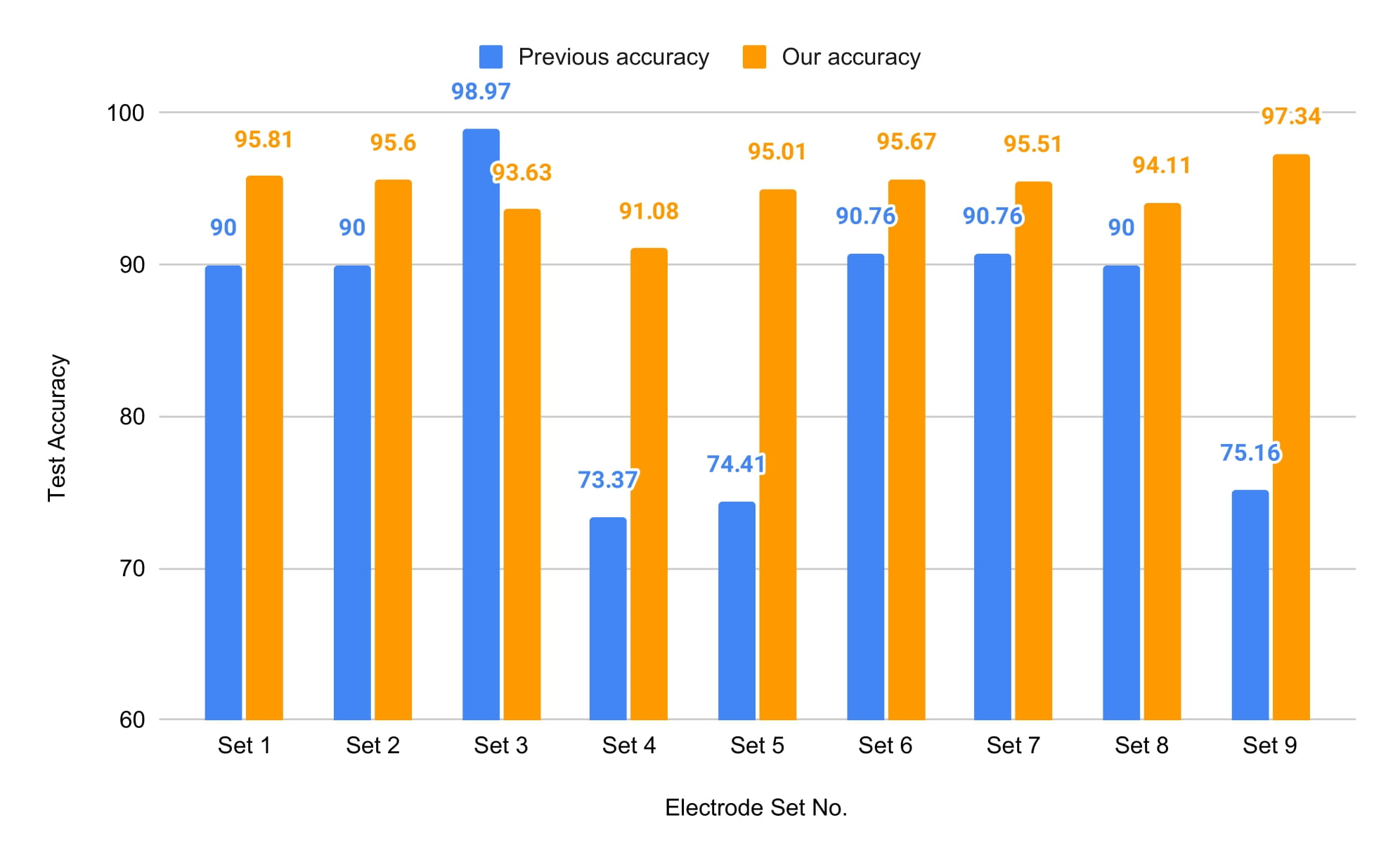}
\caption{Emotion recognition accuracy of different electrode sets} \label{Emotion recognition accuracy of different electrode set proposed by different authors}
\end{figure}

To visualize the results more clearly, the figure \ref{Emotion recognition accuracy of different electrode set proposed by different authors} is plotted with testing accuracy of our proposed model vs previous work
s accuracy with respect to electrode set number. The proposed CNN model outperforms almost all the previous work's accuracy except set three proposed by Goshvarpour et at. \cite{gosh2019}. They claim to get the high accuracy by using RSSF, Lagged Poincare Indices, and SVM. But in head to head comparison we noticed that their electrode set is not the optimal. It only got 93.63\% accuracy with our model. The electrode set with all the 32 electrodes performs the best with 97.34\% testing accuracy with the proposed model. The electrode set one by Zhang et al. using mRMR is the most optimal electrode set. With only 12 electrodes, it achieved 95.81\% testing accuracy with the proposed 1D CNN. Following that the second most optimal electrode set is number six found using ReliefF by Topic et al., which achieved 95.67\% testing accuracy with the proposed model. Zhang et al.’s ReliefF and Topic et al.’s NCA is the third and fourth most optimal electrode set with 95.60\% and 95.51\% testing accuracy respectively with the 1D CNN model.

\begin{figure}
\includegraphics[width=\textwidth]{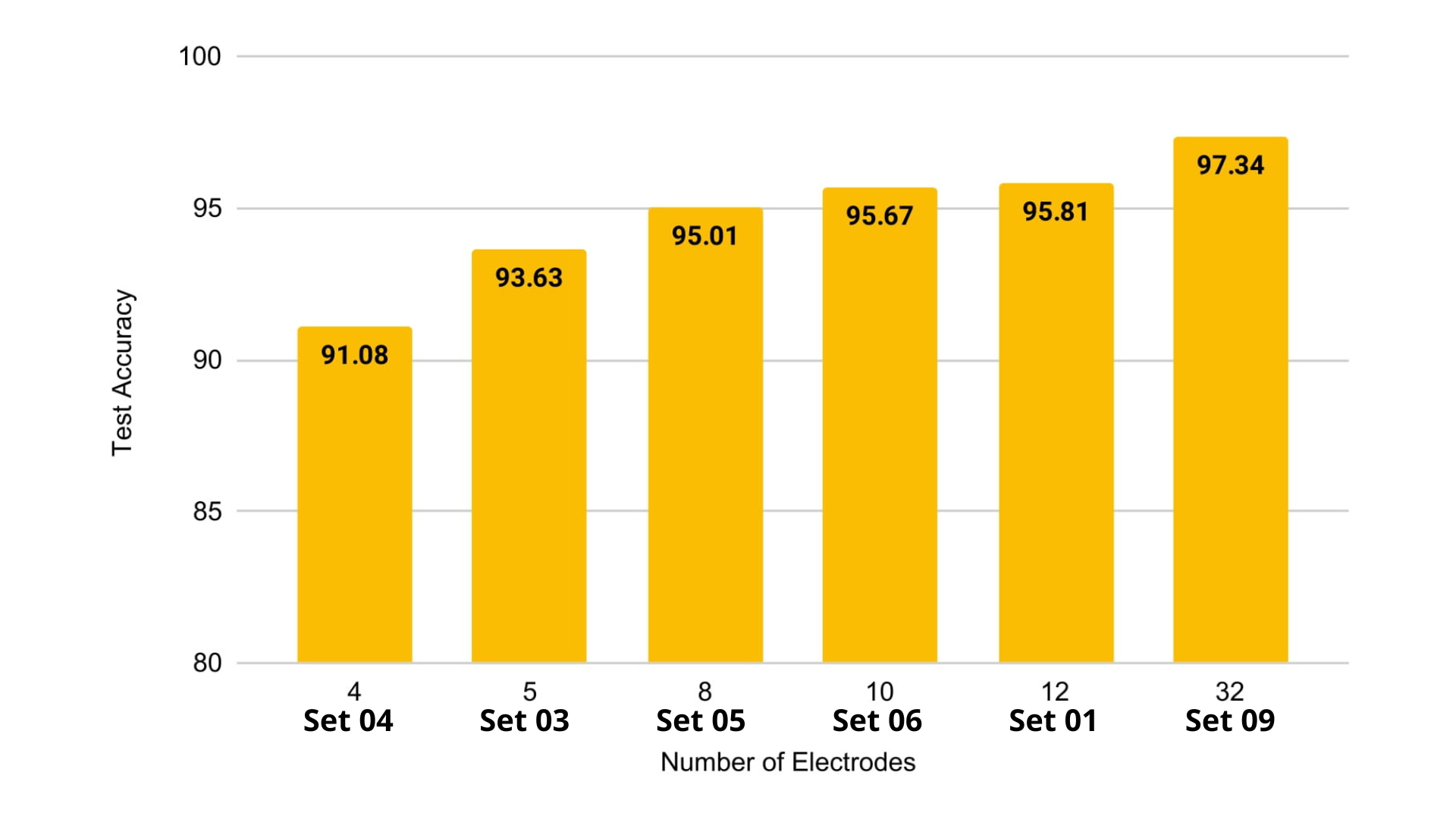}
\caption{Emotion recognition accuracy with different number of electrodes} \label{Emotion recognition accuracy with different number of electrode}
\end{figure}

To find out the correlation between the number of electrodes and testing accuracy, figure \ref{Emotion recognition accuracy with different number of electrode} is plotted. Here, electrode set 2, 6 and 7 is discarded as with the same number of electrodes there are better set which got better accuracy. It is observed that there is a clear correlation between the number of electrodes and testing accuracy. Increasing the number of electrodes from 4 to 10 greatly increases the testing accuracy. However, from the above ten electrodes, the improvement of testing accuracy is not so much. So, it can be stated that the optimal number of electrodes for emotion recognition is 10.  

\section{Conclusion}
Identifying precisely the electrodes which are optimal for recognizing emotions from brain wave is essential. This is because different portions of the brain have different roles, most of which are still unknown. In order to explore and reveal the emotion regions of brain, this study is conducted. Identifying the brain lobe which is more responsible for emotions and their respective electrodes makes it possible to reduce computational overhead and get the most optimal results.
In this study, we demonstrate that the frontal lobe is the most important brain-region for emotions. Then the evaluation of different electrode sets created by other researchers is conducted. To do the experiments, FFT is used to extract features, and a 1D-CNN with residual connection is used. The DEAP dataset was selected for this study, and the valence label was selected for all the experiments to maintain consistency. The 32 electrode set got the best testing accuracy of 97.34\%. However, the most optimal EEG electrode set is the one proposed by Zhang et al. with 12 electrodes using mRMR, which achieved  95.81\%. Those 12 electrodes are F7,P8,O1,F8,C4,T7,PO3,Fp1,Fp2,O2,P3,Fz. Also, the optimal number of electrodes is 10.
Nevertheless, there could be other electrode sets which are not yet been experimented with by any authors, thus not included in this study. Also, different electrodes are responsible for different labels of emotions like happy, sad, and angry. In future, we want to find the correlation between different labels of emotion with single electrode.

\input{references}

\end{document}

%% file: references.tex
%
%
%